\documentclass[sigconf]{acmart}

\makeatletter
\newcommand{\indep}{\protect\mathpalette{\protect\@indep}{\perp}}
\newcommand*{\@indep}[2]{\mathrel{\rlap{$#1#2$}\mkern3mu{#1#2}}}
\makeatother

\newcommand{\mme}{\mathbb{E}}
\AtBeginDocument{%
  }

\copyrightyear{2024}
\acmYear{2024}
\setcopyright{acmlicensed}
\acmConference[ICMAI 2024]{2024 9th International Conference on Mathematics and Artificial Intelligence}{May 10--12, 2024}{Beijing, China}
\acmBooktitle{2024 9th International Conference on Mathematics and Artificial Intelligence (ICMAI 2024), May 10--12, 2024, Beijing, China}
\acmDOI{10.1145/3670085.3670089}
\acmISBN{979-8-4007-1728-4/24/05}

\begin{document}

\title{A New Transformation Approach for Uplift Modeling with Binary Outcome }

\author{Kun Li}
\email{likun3@ebchinatech.com}
\affiliation{%
  \institution{Everbright Technology Co. LTD}
  \streetaddress{18 Shijingshan Rd Yi}
  \city{Shijingshan Qu}
  \state{Beijing Shi}
  \country{China}
}

\author{Liangshu Zhu}
\email{zhuliangshu@cebbank.com}
\affiliation{%
  \institution{China Everbright Bank}
  \streetaddress{25 Taipingqiao Ave}
  \city{Xicheng Qu}
  \state{Beijing Shi}
  \country{China}
}
\renewcommand{\shortauthors}{K. Li et al.}

\begin{abstract}
Uplift modeling has been used effectively in fields such as marketing and customer retention, to target
those customers who are more likely to respond due to the campaign or treatment. Essentially, it is a machine learning technique that predicts the gain from performing some action with respect to not taking it. A popular class of uplift models is the transformation approach that redefines the target variable with the original treatment indicator. These transformation approaches  only need to train and predict the difference in outcomes directly. The main drawback of these approaches is that in general it does not use the
information in the treatment indicator beyond the construction
of the transformed outcome and usually is not efficient.
In this paper, we design a novel transformed outcome for the case of the binary target variable and unlock the full value of the samples with zero outcome. From a practical perspective,
our new approach is flexible and easy to use.
Experimental results on synthetic and real-world datasets obviously show that our new approach outperforms the traditional one. 
At present, our new approach has already been applied to precision marketing in a China nation-wide financial holdings group.

\end{abstract}

\ccsdesc[500]{Computing methodologies~Machine learning}
\ccsdesc[300]{Applied computing~Marketing}
\ccsdesc[300]{Applied computing~Online shopping}

\keywords{causal learning, uplift modeling, transformation approach}

\maketitle

\section{Introduction}
Uplift modeling is a machine learning technique that predicts the gain from performing some action with respect to not taking it \cite{Gutierrez2016CausalIA, Radcliffe2007UsingCG}. Uplift modeling can be applied to a variety of real-world scenarios where the spontaneous and background outcome need to be taken into account to assess the true gains from taking an action, such as medical treatments and direct marketing campaigns \cite{ZhangWJ2022}.

Unlike standard predictive models
for outcomes in that whether or not a user is likely to convert or not, uplift modeling attempts to predict the treatment effect. It addresses this problem by using two training sets, the treatment dataset containing data on objects on which the action has been taken, and the control dataset containing data on objects left untreated. Then a model is built that predicts the difference between outcomes after treatment and without it conditional on available predictor variables \cite{Holland1986, Radcliffe1999,imbens_rubin_2015}.

A direct approach to uplift modeling is to build two separate models for treatment and control outcomes and subtract their predictions. Several estimation techniques, known as meta-learners, have been proposed in the literature. These approaches, such as the two-model approach \cite{Radcliffe2007UsingCG}, X-learner \cite{Knzel2017MetalearnersFE}, R-learner \cite{Nie2020} etc., have to train two or more models.

In many cases better results can be obtained with models which predict the difference in outcomes directly. Two major categories of estimation techniques, data preprocessing and data processing approaches, have been proposed in the literature. The first includes transformed outcome methods \cite{Jaskowski2012UpliftMF,doi:10.1073/pnas.1510489113}, which modify the outcome and extend classical machine learning techniques. The second refers to direct uplift modeling such as  Logisitc regression \cite{HANSOTIA200235,Lo2002}, KNN \cite{Alemi2009ImprovedSM,Su2012}, SVM \cite{Imai2013}, decision trees \cite{Radcliffe1999,HANSOTIA200235,5693998,GUELMAN201524}, and various neural network based methods \cite{Louizos2017CausalEI,Yoon2018GANITEEO}, which modify existing machine learning algorithms to estimate treatment effects. As ensemble methods have proven to be particularly useful in machine learing, uplift trees can be extended to more general ensemble tree models, such as causal forests \cite{doi:10.1080/01621459.2017.1319839,10.1214/18-AOS1709}. 

The advantages of the transformed outcome method are fast computation speed and easy to implement as they can leverage the existing implementation of standard classification
and regression models. The main drawback of these approaches
is that in general it does not use the information in
the treatment indicator beyond the construction of
the transformed outcome and usually is not efficient. 
Especially, the binary outcome case with low response rate is a challenge for these type approaches.

In order to take advantage of the transformed outcome method and improve its performance in binary outcome case, we propose a new transformation approach for uplift modeling with binary outcome. 
Our proposed approach modifies the transform formula in \cite{doi:10.1073/pnas.1510489113} and still satisfies the property of the transformed target's expectation. 
It is demonstrated through empirical experiments that our new approach can achieve better uplift estimation compared with the existing models. Extensive experimental results on synthetic data show the success of our new approach in uplift modeling. We also train different approaches and evaluate the performance on a large real-world marketing campaign data set from a national financial holdings group in China. As expected, the corresponding results show a significant improvement in the evaluation metrics in terms of the different metrics.

The remainder of this manuscript is structured as follows. First, in Sect. \ref{sec:umtof}, 
problem formulation, existing transformation approaches for the uplift model and our new approach are provided. 
Then Sect. \ref{sec:er} gives the details of our experimental results in the synthetic data set and the real world marketing campaign, and finds that our method performs well in a wide variety of settings relative to the old approaches. Finally, the conclusions are presented in Sect. \ref{sec:con}.

\section{Uplift Model and Transform Outcome Method}\label{sec:umtof}
For marketing campaigns, uplift modeling is different from standard predictive models
in marketing in that it attempts to predict the effect of treatment
rather than whether or not a user is likely to convert or
not. These two are not necessarily the same, because a high treatment
effect and a high probability to convert do not necessarily
coincide.

\subsection{Problem Definition}\label{subsection:2_1}
In a marketing campaign, a customer $i$ is either in the treatment group or in the control group.  Let $W_i \in \{0, 1\}$ be the binary indicator for the treatment/control group. $W_i = 0$ indicates that customer $i$ has not been targeted and receives the control treatment. $W_i = 1$
indicates that customer $i$ has not been targeted and receives the active treatment. 

Let $Y$ be the target variable. Following the notation in \cite{doi:10.1073/pnas.1510489113}, the realized outcome for customer $i$ is the potential outcome corresponding to the treatment received.
\begin{align}\label{Y_obs}
Y_i^{obs}=Y(W_i)=\left\{
\begin{array}{cc}
     Y_i(0)& \mathbf{if}\, W_i=0 \\
     Y_i(1)& \mathbf{if}\, W_i=1
\end{array}
\right.
\end{align}
i.e. \[Y_i^{obs}=W_iY_i(1)+(1-W_i)Y_i(0).\]
Let $X_i$ be a vector of components of $K$ of features, covariates, or pretreatment variables, known not to be affected by the treatment. The data consists  $(Y_i^{obs},W_i,X_i)$.

The assumption of randomization condition on the unconfoundedness was introduced in \cite{10.1093/biomet/70.1.41} as follows
\begin{equation}\label{eq:unconf}
 (Y_i(1),Y_i(0))\indep W_i\mid x_i,
\end{equation}
which means treatment assignment $W_i$ and response $ (Y_i(1),Y_i(0))$ are known to be conditionally independent with given $x_i$.

Let $p(x) = P(W_i = 1\mid X_i = x)$ be the
probability of conditional treatment (the 'propensity score' defined by \cite{10.1093/biomet/70.1.41}). In a randomized
experiment with constant treatment assignment probabilities $p(x) = p$ for all values of $x$. Let
$p = P(W_i = 1)$ be the marginal treatment probability.

The conditional average treatment effect (CATE) is defined as
$\tau(x):=\mme \left[Y_i(1)-Y_i(0)\mid X_i = x\right]$. 

It is called "uplift" in uplift modeling, and it is the difference in behavior. A large part of the uplift modeling literature (e.g., \cite{doi:10.1089/big.2017.0104,doi:10.1073/pnas.1510489113}) is focused on obtaining accurate estimates of and inferences for the conditional average treatment effect $\tau(x)$.

A lot of different approaches, including tree-based and regression-based methods,
are adopted for developing uplift models.
These approaches can be grouped into data
preprocessing and data processing approaches. Whereas
data preprocessing approaches essentially adopt traditional
predictive analytics in an adapted setup for learning
an uplift model, the data processing approaches concern
adapted predictive analytics for developing uplift models.

\subsection{Existing Transformation Approaches}
Transformation approaches that redefine
the target variable,
are  widely popular in science
and industry, given the predictive strength and interpretability
of the resulting model.

A classical variable transformation for the binary outcome was proposed
by \cite{Jaskowski2012UpliftMF}. The target variable
is transformed by defining a new target variable
$Z\in\{0,1\}$ as follows:
\begin{equation}
Z_i=\left\{
\begin{array}{cc}
     1& \mathbf{if}\, W_i=1,Y_i=1\,\mathbf{or} \,W_i=0,Y_i=0\\
     0& \mathbf{if}\, W_i=1,Y_i=0\,\mathbf{or} \,W_i=0,Y_i=1
\end{array}
\right.
\end{equation}
With the assumption $p(W_i=1)=p(W_i=0)=1/2$, the resulting model
then allows estimating uplift as the difference between
the success probabilities in treatment and control
groups 
\begin{align}
&P(Y=1\mid X=x,W=1)-P(Y=1\mid X=x,W=0)\nonumber\\
=&2P(Z=1\mid X=x)-1
\end{align}
Due to this classical transformation,  the uplift modeling problem is
converted into a binary classification problem with label $Z$, allowing to adopt any traditional classification technique. However it usually fails to estimation the correct uplift effect when the response rate of $Y_i=1$ is extremely low, which is common in a  online marketing campaign in real world.

Another transformation
\begin{equation}\label{eq:trans2}
Z^*_i =  Y_i\frac{W_i-p(X_i)}{p(X_i)(1-p(X_i))},
\end{equation}
is suggested by \cite{doi:10.1073/pnas.1510489113}. 
It works for both binary and continuous target variables. It also satisfies 
\begin{equation}\label{exp_tau}
\mme \left[Z^*_i\mid X_i = x\right] = \tau(x)
\end{equation}
with the unconfounded assumption \eqref{eq:unconf} and does not suffer from low response rate in binary outcome case. All the machine learning methods for regression can be adopted for this model, which is the key attraction of this transformation approach.

\begin{table}
\centering
\begin{tabular}{ll|lc|c}
\hline
$W$  &$Y$   &$Z$  &$Z^*$&\# of samples \\
\hline
1      &1     &   1   &  $\frac{1}{p}$&$pN(r+\Delta r)$       \\
0      &0     &   1   &  0 & $(1-p)N(1-r)$     \\
1      &0     &   0   & 0&  $pN(1-r-\Delta r)$   \\
0      &1     &   0   &  $-\frac{1}{1-p}$  & $(1-p)Nr$    \\
\hline
\end{tabular}
\caption{Comparison of two transformation approaches. Here $r=P(Y_i=1\mid W_i=0)$ and $r+\Delta r=P(Y_i=1\mid W_i=1)$}
\label{tab:comp1}
\end{table}

\subsection{Response rate based modification for binary target variable}

For the binary target variable case $Y_i$, the main drawback of the transformed result $Z^*$ is that, in general, it
is not efficient because it does not use the information in the treatment indicator for the samples with $Y_i=0$. According to Table \ref{tab:comp1}, it is obvious that $Z_i^*=0$ for $Y_i=0$ regardless of whether it is treated or not.

The goal of this paper is to study how we can design a new approach which still satisfies the property \eqref{exp_tau} and distinguishes all the different cases of $W$ and $Y$. As motivation for our approach, note that 
$\mme[(W_i-p(X_i))/(p(X_i)(1-p(X_i)))|X_i = x] = 0$ and only the samples with $Y_i=0$ misses the information of $W_i$. 
A natural modification of the transformation approach in \cite{doi:10.1073/pnas.1510489113} can be proposed as follows
\begin{equation}\label{tran_new}
\hat{Z}^*_i = (Y_i-C)\frac{W_i-p(X_i)}{p(X_i)(1-p(X_i))},
\end{equation}
here $C$ is a constant. 
The new transformed outcome $\hat{Z}^*$ also has the following proposition.
\begin{proposition}
Suppose that Assumption \eqref{eq:unconf} holds. Then:
\[ \mme\left[ \hat{Z}^*_i\mid X_i=x\right]=
\tau(x).\]
\end{proposition}
\noindent{\bf Proof:}
Expand on this equality by definition,
\[ \mme\left[ \hat{Z}^*_i \mid X_i=x\right]=
\mme\left[ \left.(Y^{obs}_i-C)\frac{W_i-p(X_i)}{p(X_i) (1-p(X_i))}\right| X_i=x\right]
\]
\[ =
\mme\left[ \left.\left(W_i Y^{obs}_i +
(1-W_i) Y^{obs}_i-C\right) \frac{W_i-p(X_i)}{p(X_i) (1-p(X_i))}
\right| X_i=x\right].
\]
Because $W_i Y^{obs}_i=W_i Y_i(1)$ and 
$(1-W_i) Y^{obs}_i=(1-W_i) Y_i(0)$,
 this can be rewritten as
 
\begin{align*}
 &\mme\left[\left. \left(W_i Y_i(1) +
(1-W_i) Y_i(0) -C\right)\frac{W_i-p(X_i)}{p(X_i) (1-p(X_i))}
\right|X_i=x\right]\\
&=
\mme\left[\left.  Y_i(1) \frac{W_i (1-p(X_i))}{p(X_i) (1-p(X_i))}\right|X_i=x\right]\\
&-\mme\left[\left. 
 Y_i(0) \frac{(1-W_i) p(X_i)}{p(X_i) (1-p(X_i))}
\right|X_i=x\right]\\
&\quad
+C\mme\left[\left. 
\frac{W_i- p(X_i)}{p(X_i) (1-p(X_i))}
\right|X_i=x\right]\\
&=
\mme\left[\left.Y_i(1) W_i\right|X_i=x\right]\frac{1}{p(X_i)}
-\mme\left[\left. 
 Y_i(0)(1-W_i)
\right|X_i=x\right]\frac{1}{1-p(X_i)}\\
&\quad+C\left(\mme\left[\left.  W_i=1\right|X_i=x\right] 
\frac{1}{p(X_i)}-\mme\left[\left.  W_i=0\right|X_i=x\right] 
\frac{1}{1-p(X_i)}\right).
\end{align*}
Because of unconfoundedness \eqref{eq:unconf} this equals to
\begin{align*}
&\mme\left[\left. \hat{Z}^*_i\right|X_i=x\right]\\
=&
\mme\left[\left.  Y_i(1)\right|X_i=x\right]
\mme\left[\left.  W_i\right|X_i=x\right] 
\frac{1}{p(X_i)}\\
-&\mme\left[\left. 
 Y_i(0)
\right|X_i=x\right]
\mme\left[\left. 
1-W_i\right|X_i=x\right] 
\frac{1}{1-p(X_i)}\\
+&C\left(\frac{p(X_i)}{p(X_i)}-\frac{1-p(X_i)}{1-p(X_i)}\right)\\
=&\mme \left[Y_i(1)\mid X_i = x\right]-\mme\left[Y_i(0)\mid X_i = x\right]=\tau(x).
\end{align*}
$\square$

The values of $\hat{Z}^*$ for the binary target variable $Y_i$ are shown in Table \ref{tab:comp2}. The fact that the samples with $(W,Y)=(1,1)$ or $(0,0)$ should be more valuable for treatment than the ones with $(W,Y)=(1,0)$ or $(0,1)$, indicates that $C$ have to lay in the interval $[0,1]$, i.e. $0<C<1$. In addition, the treatment effect
of the samples with different $(W,Y)$ should depend on the response rate of $Y_i$ (i.e. $P(Y_i=1)$). When the response rate is very small, e.g. 

$r=pr(Y_i=1|W_i=0) \ll 1$, the samples with $(W,Y)=(1,1)$ should be more valuable than the samples with $(W,Y)=(0,0)$.

\begin{table}
\centering
\begin{tabular}{ll|cc}
\hline
$W$  &$Y$   &$Z^*$  &$\hat{Z}^*$ \\
\hline
1      &1     & $1/p$ &$(1-C)/p$      \\
0      &0     & 0 & $C/(1-p)$     \\
1      &0     & 0& $-C/p$    \\
0      &1     & $-1/(1-p)$& $-(1-C)/(1-p)$    \\
\hline
\end{tabular}
\caption{Comparison of two transformation approaches.}
\label{tab:comp2}
\end{table}

\section{Approach evaluation on both synthetic and real world data}\label{sec:er}
In this section, the effectiveness of our new approach is evaluated on synthetic and real world datasets, respectively. Since our approach is designed as a modification of the transformed outcome $Z^*$ in \cite{doi:10.1073/pnas.1510489113},
we implement it and compare it with $Z^*$, which are referred to as baseline.

\subsection{Experiments on Synthetic Dataset}

\textbf{Dataset} We can test the methodology with numerical simulations. That is, generating synthetic datasets with known causal and noncausal relationships between the outcome, action (treatment/control), and some confounding variables. More specifically, both the outcome and the action/treatment variables are binary. A synthetic dataset is generated with the $make\_uplift\_classification$ function in the "Causal ML" package, based on the algorithm in \cite{Guyon2003DesignOE}.
There are 10,000 instances for the treatment and control groups, with different response rates (high and low response rate, seen in Table \ref{tab:data}) respectively.
In the high response rate case, the treatment group and a control group have 5021 and 3038 positive outcome $Y=1$ samples with response rates of 50.21\% and 30.38\%. In the low response rate case, the treatment group and a control group have 784 and 553 positive outcome $Y=1$ samples with response rates of 7.84\% and 5.53\%. The correlation coefficient of $Z$ and $W$ is -0.198 in the high response case, compared with -0.867 in the low response case.   

The input consist of 5 features in three categories. 3 of them are used for base classification, which are composed of 2 informative and 1 irrelevant variables. 2 positive uplift variables are created to testify positive treatment effect. 

\begin{table}
\centering
\begin{tabular}{ll|ll|c|c}
\hline
&  &\# of   &\# of   & &\\
&$W$  &$Y=1$   &$Y=0$  &RR&$\rho(Z,W)$ \\
\hline
High&0      &3038     & 6962 & 30.38\% &-0.198 \\
RR&1      &5021     & 4979 & 50.21\%  &\\
\hline
Low&0      &553     & 9447 & 5.53\%  &-0.867\\
RR&1      &784     & 9216 & 7.84\%  &\\
\hline
\end{tabular}
\caption{Statistics Information of Two Synthetic Datasets (high and low response rate(RR)).}
\label{tab:data}
\end{table}

\begin{figure*}[htbp]
\centering
\includegraphics[width=0.49\textwidth]{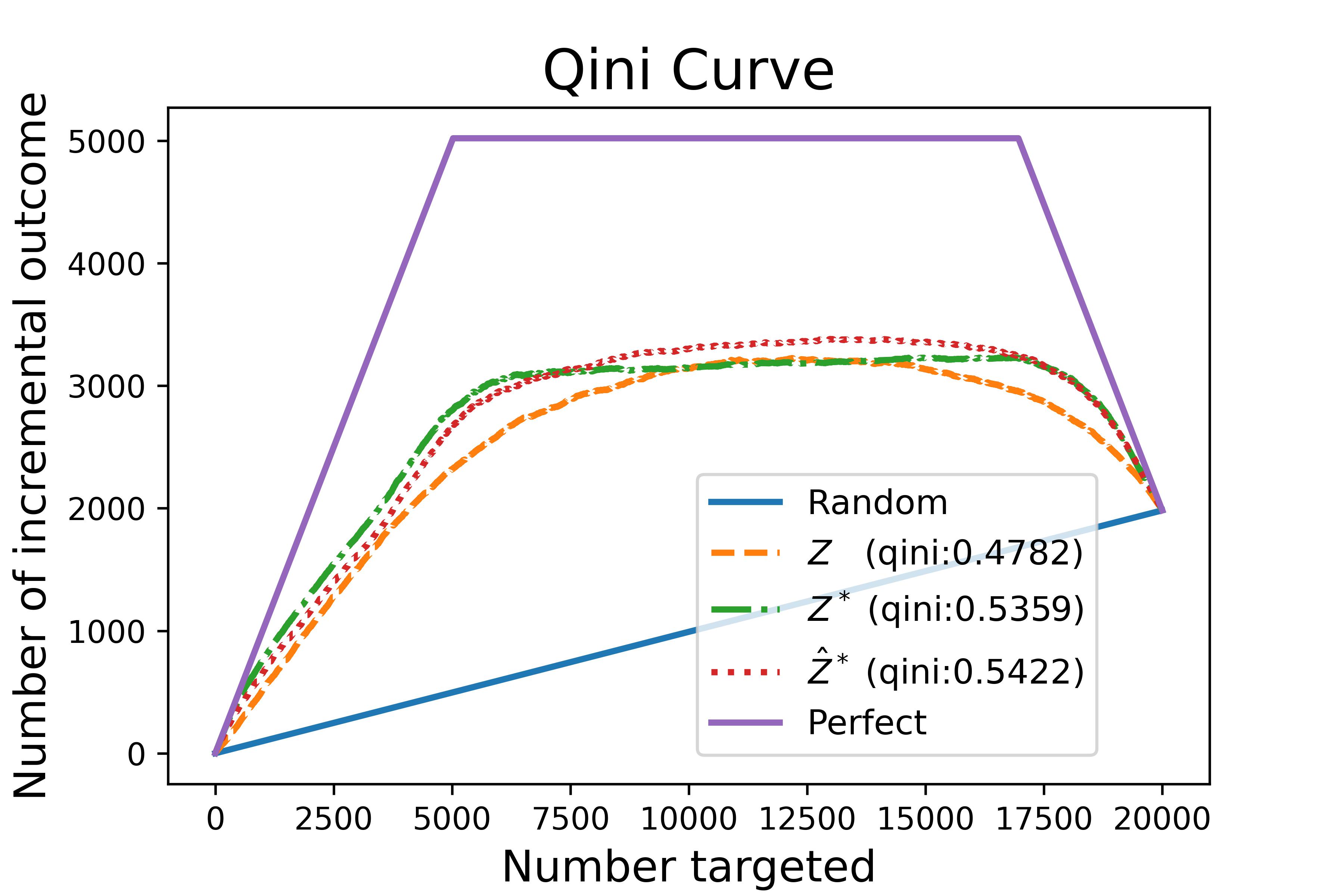}
\includegraphics[width=0.49\textwidth]{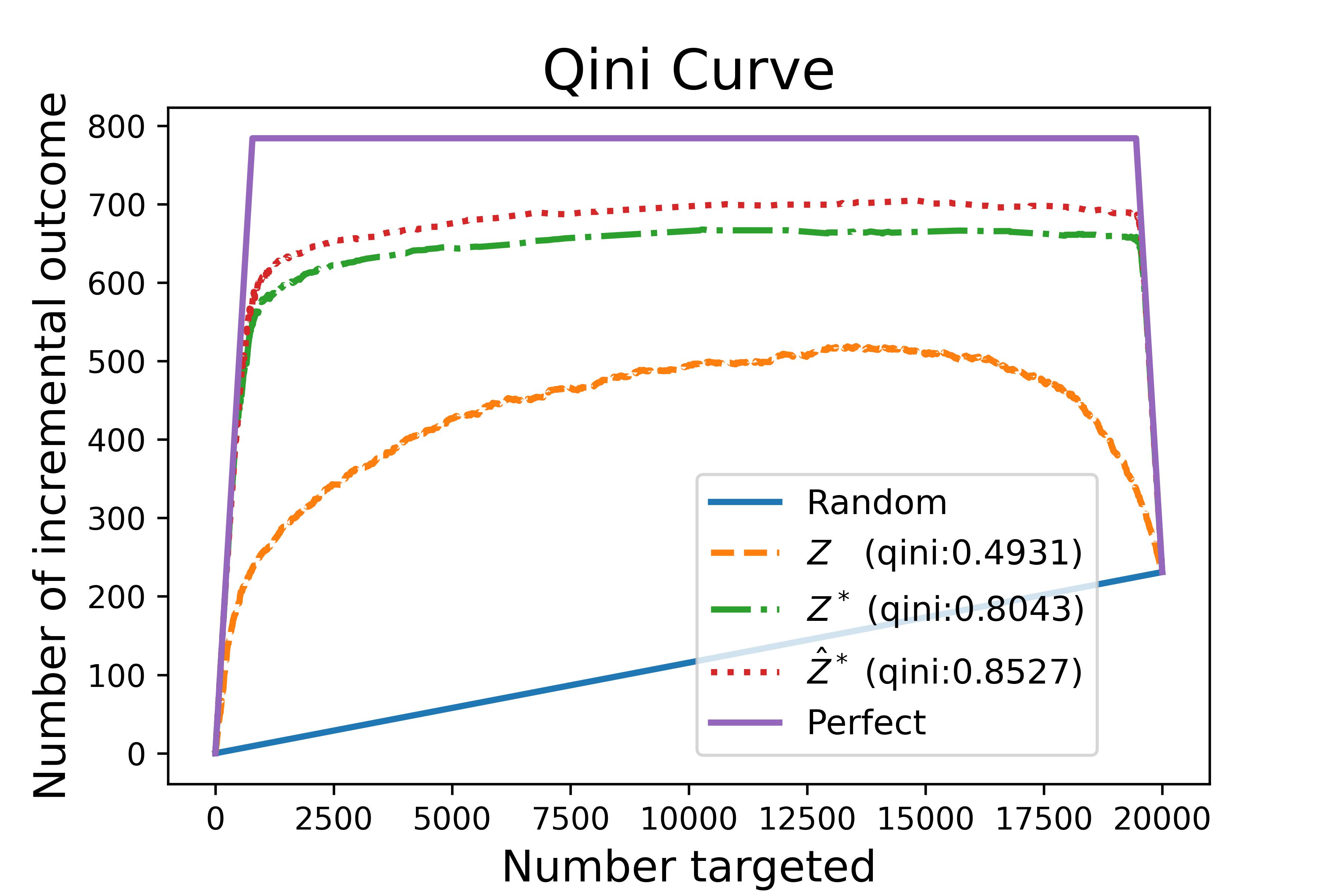}
\caption{Model performance in Qini Curve of different approaches ($Z$, $Z^*$ and $\hat{Z}^*$) for synthetic dataset with high (LEFT) and low (RIGHT) response rate.}
\label{fig:low_high_rate} 
\end{figure*}

\textbf{Parameters and Results} An eXtreme Gradient Boosting 
(XGBoost) algorithm for regression with same parameters is adapted as the machine learning model for different transformation approaches  $Z^*$ and $\hat{Z}^*$.  In addition,
an eXtreme Gradient Boosting (XGBoost) algorithm for classification with same maximum tree depth for base learners and number of gradient boosted trees is adapted as the machine learning model for the classical transformation approach $Z$.

Most of the uplift literature resort to aggregated
measures such as uplift bins or curves. Two key metrics involved are the area under the qini curve and the coefficient\cite{Gutierrez2016CausalIA}, respectively. 
The higher these values, the better the uplift model. 

As shown in Figure \ref{fig:low_high_rate}, qini coefficient increases from 0.5359 at baseline $Z^*$ to 0.5422 at $\hat{Z}^*$ with 
0.4782 at $Z$ for high response rate dataset. For the low response rate data set, the qini coefficient increases from 0.8043 at baseline $Z^*$ to 0.8527 at  $\hat{Z}^*$ with 0.4931 at $Z$. In this case, the classical transform $Z$ has a stronger correlation with $W$, which should indicate well-randomized groups. It cause a significant downgrade in the performance of the model for $Z$ with a comparison with the ones of $Z^*$ and $\hat{Z}^*$.
It is obvious that the performance of the new approach $\hat{Z}^*$ in different response rate cases is remarkably improved, especially in the low response rate case.


\subsection{Experimental results in a marketing campaign}
\textbf{Dataset} We further extend the new approach to precision marketing for new customer application. A telephone marketing campaign via AI chatbots is designed to promote customers to buy products from a national financial holdings group in China. The target outcome is 1 / 0, indicating whether or not a customer would buy the given product. 
The data shown in Table \ref{tab:data2} contains 106,512 individuals, consisting of a treated group of 26,470 (chatting with AI chatbots) and a control group of 80,042 (not chatting with AI chatbots). These two groups have 261 and 480 product buyer with response rates of 0.986\% and 0.600\%, which are typical values in real world marketing practice. There are 20 variables in all, which are characterized as asset under management and credit card information, customer demographics et al.

\begin{table}
\centering
\begin{tabular}{l|ll|c|c}
\hline
  &\# of   &\# of   & &\\
$W$  &$Y=0$   &$Y=1$  &Response Rate&P(W) \\
\hline
0      &79462     & 480 & 0.5997\% &75.148\%\\
1      &26209     & 261 & 0.9860\% &24.852\%\\
\hline
\end{tabular}
\caption{Statistics Information of A  Marketing
Campaign Dataset.}
\label{tab:data2}
\end{table}

\begin{figure*}[htbp]
\centering
\includegraphics[width=0.49\textwidth]{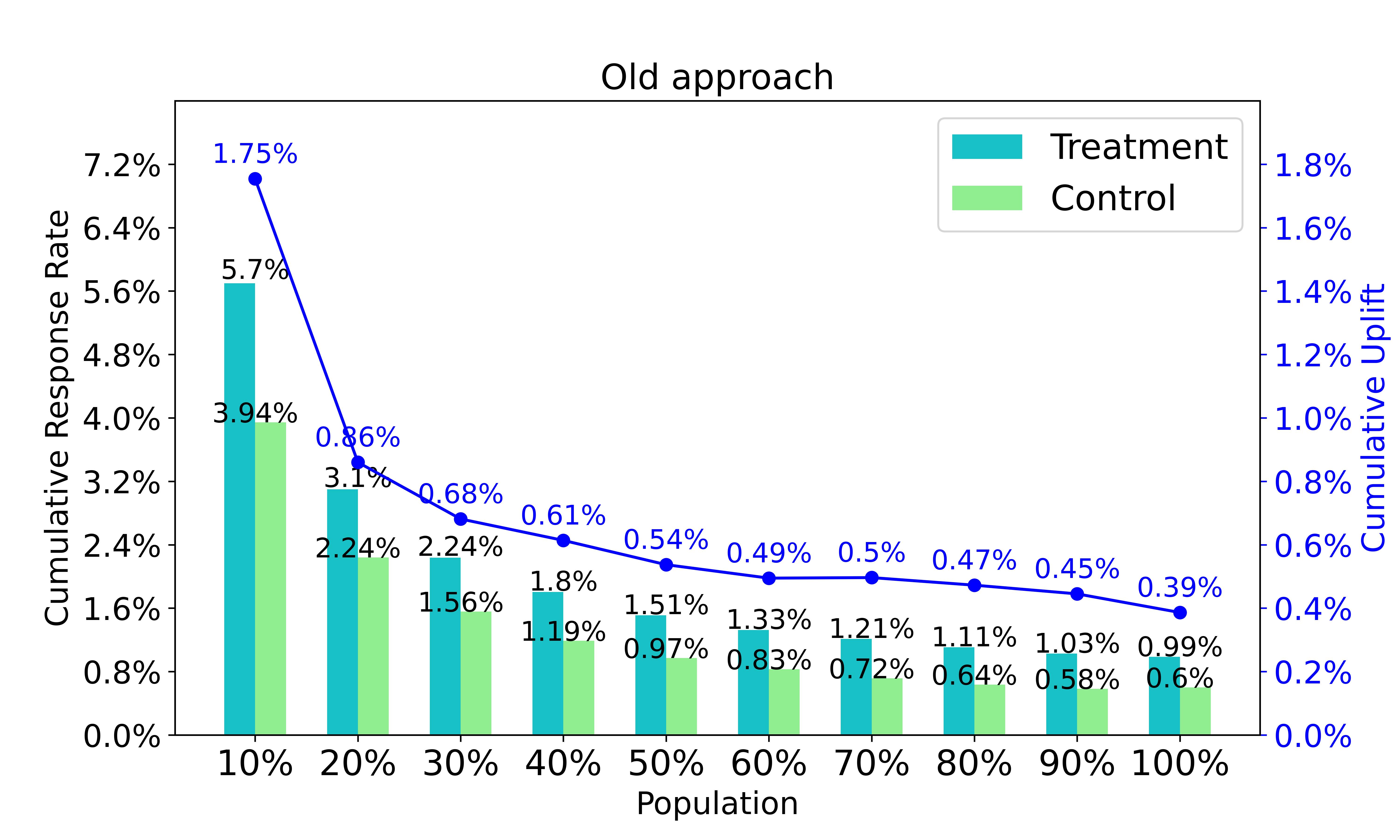}
\includegraphics[width=0.49\textwidth]{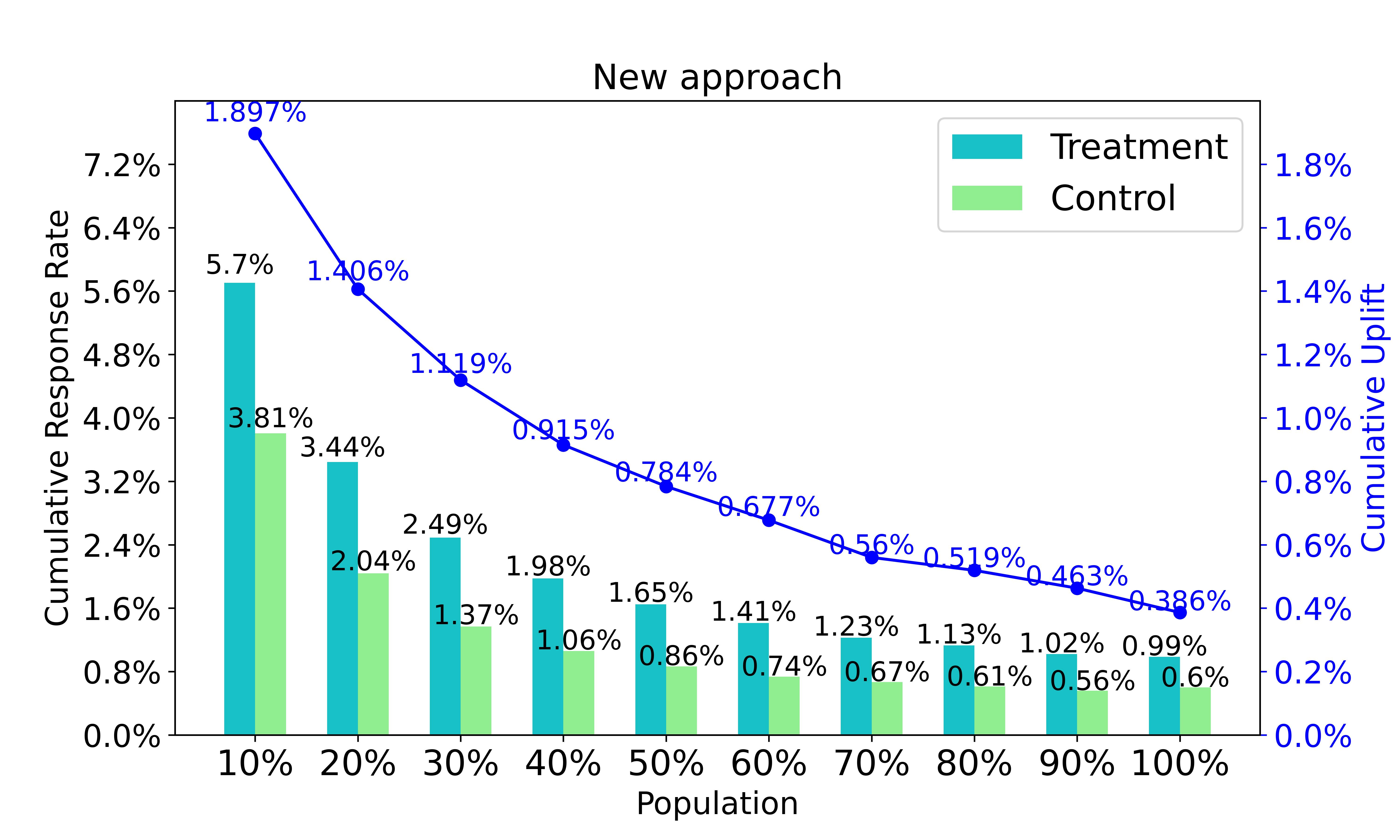}
\caption{Model performance in cumulative uplift of old approach $Z^*$ (LEFT) and new approach $\hat{Z}^*$ (RIGHT) for a real world marketing campaign.}
\label{fig:realworld} 
\end{figure*}

\textbf{Parameters and Results} We still choose an eXtreme Gradient Boosting (XGBoost) algorithm with same parameters as the machine learning model for different approaches  $Z^*$ and $\hat{Z}^*$. 
Beside the area under uplift curve (AUUC) and Qini curve, the cumulative uplift at different percentiles is investigated. 
In marketing practice, the higher uplift at top K percentile, the better model performance. 

\begin{table}[htbp]
\centering
\begin{tabular}{llll}
\hline
Model  &AUUC   &Qini \\
\hline
Old approach $Z^*$    &0.001949     & 0.1268      \\
New approach $\hat{Z}^*$  &0.002852    & 0.1526       \\
\hline
\end{tabular}
\caption{Model performance of old approach $Z^*$ and new approach $\hat{Z}^*$ for a real world marketing campaign.}
\label{tab:realworld}
\end{table}

According to Table \ref{tab:realworld}, 
it is obvious that the area under uplift curve (AUUC) and Qini curve is remarkably improved due to new transform. 
Compared to 0.001949 in the old approach $Z^*$, the AUUC improves significantly to 0.002852 using our new approach $\hat{Z}^*$. 
At the same time, similar performance is clearly shown for the area under Qini curve.

Furthermore, we calculate the cumulative uplift values at each 10 percentile. 
It can be seen from Figure \ref{fig:realworld} that the uplift values with the new approach $\hat{Z}^*$ are better than those with the old approach $Z^*$. Especially, the improvements at the top 20th, 30th, 40th, 50th, and 60th percentile are all over 30\%. 
That is to say, our new approach $\hat{Z}^*$ can distinguish the target samples from the treatment effectively.

\section{Conclusion}\label{sec:con}
This manuscript proposes a new transformation approach for uplift modeling with binary outcome, which is the most common case in precision marketing. 
Differently from the traditional transformation approach, the new approach involves the information of treatment even when the outcome is zero. With the help of our new approach the model performance over different statistics metrics on both synthetic and real world marketing campaign data is highly improved.  

\begin{acks}
The authors appreciate the colleagues at China Everbright Group for our valuable discussions on business understanding and inspiration for the application design. 
\end{acks}


\bibliographystyle{ACM-Reference-Format}
\bibliography{example_paper}



\end{document}